\documentclass{article}
\usepackage[utf8]{inputenc}
\usepackage{authblk}
\usepackage{setspace}
\usepackage[margin=1.25in]{geometry}
\usepackage{graphicx}
\graphicspath{ {./figures/} }
\usepackage{subcaption}
\usepackage{amsmath}
\usepackage{amssymb}
\usepackage{float}
\usepackage{indentfirst}

\usepackage[super,square,comma,numbers,sort&compress]{natbib}

\title{Mechanical Analysis of Extraction and Straightening of Parachute Suspension Lines with Binding Tapes Using Physics-Informed Neural Networks }

\author[1*$\dag$]{Xiang Zhao}
\author[2$\dag$]{Ronghui Quan}

\affil[1]{Department of Systems Engineering, Nanjing University of Aeronautics and Astronautics, Nanjing, China.}
\affil[*]{sz2415099@nuaa.edu.cn(X. Zhao), quanrh@nuaa.edu.cn(R. Quan)}

\date{}

\usepackage{hyperref}

\title{Mechanical Analysis of Parachute Suspension Line Deployment with Binding Tapes Using PINN}

\author[1$\dag$]{Xiang Zhao}
\author[1*]{Ronghui Quan}
\author[1]{Yaqi Xiao}
\author[2]{Junlin Chen}
\affil[1]{Nanjing University of Aeronautics and Astronautics, Nanjing, China.}
\affil[2]{Beijing Institute of Space Mechanics and Electricity,Beijing, China.}
\affil[*]{quanrh@nuaa.edu.cn(R. Quan)}

\date{}

\usepackage{hyperref}

\begin{document}
\maketitle

\begin{abstract}
Parachutes are widely utilized in aviation, aerospace and lifesaving missions. As the initial stage of parachute deployment, suspension line extraction and straightening directly determines the smooth implementation of subsequent inflation procedures. This ultra-short process involves intricate dynamic load variations. Most existing studies adopt numerical integration of ordinary differential equations to calculate line tension, yet this method fails to rapidly acquire tension values at arbitrary positions along suspension lines. This paper develops a physics-informed neural network (PINN) algorithm for tension prediction during line extraction and straightening, which outperforms traditional integration methods in both computational efficiency and numerical accuracy. Furthermore, the regulatory law of binding tape parameters on line dynamic tension is investigated. Comparative validations against flight test data and conventional numerical results verify the reliability and effectiveness of the proposed PINN framework. 
\end{abstract}

\section{Introduction}

Parachutes are widely applied in aviation, aerospace, lifesaving and other fields\cite{Ref1}. Their core functions include decelerating loads, maintaining stability, and ensuring the safe landing of personnel and equipment. The parachute deployment process is critical, as it directly determines the overall safety of the system \cite{Ref2,Ref3,Ref4}. As a key component, suspension lines support the inflation of the canopy and transmit loads as well as aerodynamic drag. The line extraction and straightening process refers to the entire sequence from the suspension lines being pulled out and gradually tensioned to the full opening of the main canopy. This process features an extremely short duration and rapid state changes, accompanied by complex force variations \cite{Ref5,Ref6,Ref7,Ref8}. Current studies fail to analyze the full-range force characteristics of suspension lines from extraction to full main canopy opening for parachute systems equipped with binding tapes and various types of parachutes \cite{Ref9,Ref10,Ref11}. This deficiency results in insufficient comprehensive theoretical guidance for suspension line design, which may lead to line breakage and canopy damage, and further impair the safety and reliability of parachutes \cite{Ref12,Ref13,Ref14,Ref15}.

Scholars at home and abroad have mainly focused on canopy inflation and overall deceleration performance in parachute research. In 1984, Purvis established the LINESAIL simulation model by simplifying the parachute-load system into lumped mass nodes connected by springs. This model took aerodynamic force, elastic force, line tension and pack friction of extracted lines into account, and conducted preliminary research on the line sail phenomenon during line straightening. Aiming at the deployment mode with prior line extraction, Earle K. Huckins III proposed a novel snatch load prediction method considering elastic wave propagation. Although existing mechanical models and flow field simulation methods have been adopted to investigate the performance of fully deployed parachutes, they rarely cover the entire line extraction process and lack in-depth analysis on the constraint effect of binding tapes. For a long time, the dynamic analysis of suspension lines in engineering practice mainly relies on numerical integration of ordinary differential equations (ODE) and the finite element method (FEM)\cite{Ref16,Ref17,Ref18,Ref19}. In recent years, the lumped mass-spring-damper (MSD) discrete model and Physics-Informed Neural Network (PINN) have attracted increasing attention\cite{Ref37,Ref38,Ref39}. Overall, existing researches have prominent limitations: they cannot systematically analyze the force evolution of suspension lines throughout the whole deployment process, and the constraint effect of binding tapes is also poorly studied.

In this paper, the suspension line is divided into multiple segments, each simplified as a spring-damper unit with lumped mass at both ends. The motion of nodes on extracted line segments is governed by aerodynamic force and line tension. The mechanical properties of the entire suspension line are derived based on the force analysis of a single particle. With the influence of the quantity and layout of binding tapes on line tension fully considered, numerical calculations are carried out to analyze the force variation of suspension lines from being pulled out of the pack to full straightening. The research findings can provide references for subsequent research and optimal design of parachute suspension lines. As an innovative research paradigm, PINN can learn continuous solutions of partial differential equations and realize interpolation and prediction at arbitrary spatial and temporal points, offering a new approach to solve similar engineering problems \cite{Ref40}. 

\section{Materials and Methods}

This section first constructs the force field of a single particle on the suspension line, and then establishes a continuous extraction model for the suspension line. The developed PINN model is presented. On the basis of the previously completed tension calculation model for particles on the suspension line, a new tension calculation model considering the effect of binding tapes is further established. 

\subsection{Force Model of Suspension Line Particles }

This chapter starts with a force analysis of a single particle on the suspension line. The multi-particle spring-damper model is a widely adopted mechanical analysis method\cite{Ref20,Ref21,Ref22,Ref23}. As shown in Figure 1a, the suspension line is discretized into a series of particles interconnected by springs and dampers \cite{Ref24,Ref25,Ref26}. The springs simulate the elastic deformation of the suspension line, while the dampers represent damping forces generated during line movement\cite{Ref27,Ref28}. This model can intuitively characterize the force variation of the suspension line and is applicable for analyzing the complex dynamic behavior during line extraction. Nevertheless, it suffers from long computation time, and its accuracy is dependent on the number of discrete particles.  

\begin{figure}[H]
    \centering
    \begin{subfigure}{0.65\textwidth}
        \includegraphics[width=\textwidth]{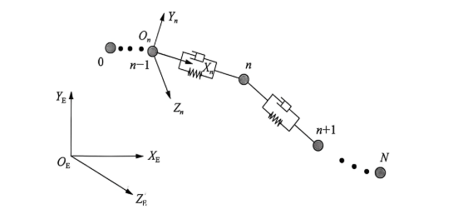}
        \caption{}
        \label{fig:1a}
    \end{subfigure}
    \hfill
    \begin{subfigure}{0.35\textwidth}
        \includegraphics[width=\textwidth]{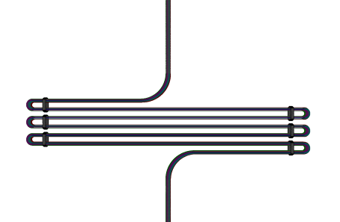}
        \caption{}
        \label{fig:1b}
    \end{subfigure}
    \caption{(\subref{fig:1a}) Mass-spring-damper model of line segments. (\subref{fig:1b}) Schematic diagram of suspension lines inside the parachute pack.}
    \label{fig:1}
\end{figure}
As illustrated in Figure 1b , the suspension line is stowed in a Z-shaped configuration inside the parachute pack. The arc length of the folded section is defined as \(s \in [0.20,0.81]\). By alternately folding the line at an angle of 180°, the 14.64 m-long suspension line is compacted to a vertical height of approximately 0.2 m. The straight segment above the folded region corresponds to \(s \in [0.81,1.0]\) and is permanently connected to the pilot parachute. The straight segment below the folded region is denoted as \(s \in [0.0,0.20]\) and remains attached to the payload. The entire suspension line along its central axis is divided into 100 discrete particles. 

The gravity, spring,damping ,aerodynamic and friction force acting on each line segment are calculated as the average values of the corresponding forces on adjacent particles, as presented in Figure 2a and Figure 2b. For a particle located at arc length s, the resultant force consists of four components, which are described as follows: 
\begin{figure}[H]
    \centering
    \begin{subfigure}{0.4\textwidth}
        \includegraphics[width=0.9\textwidth, height=2in]{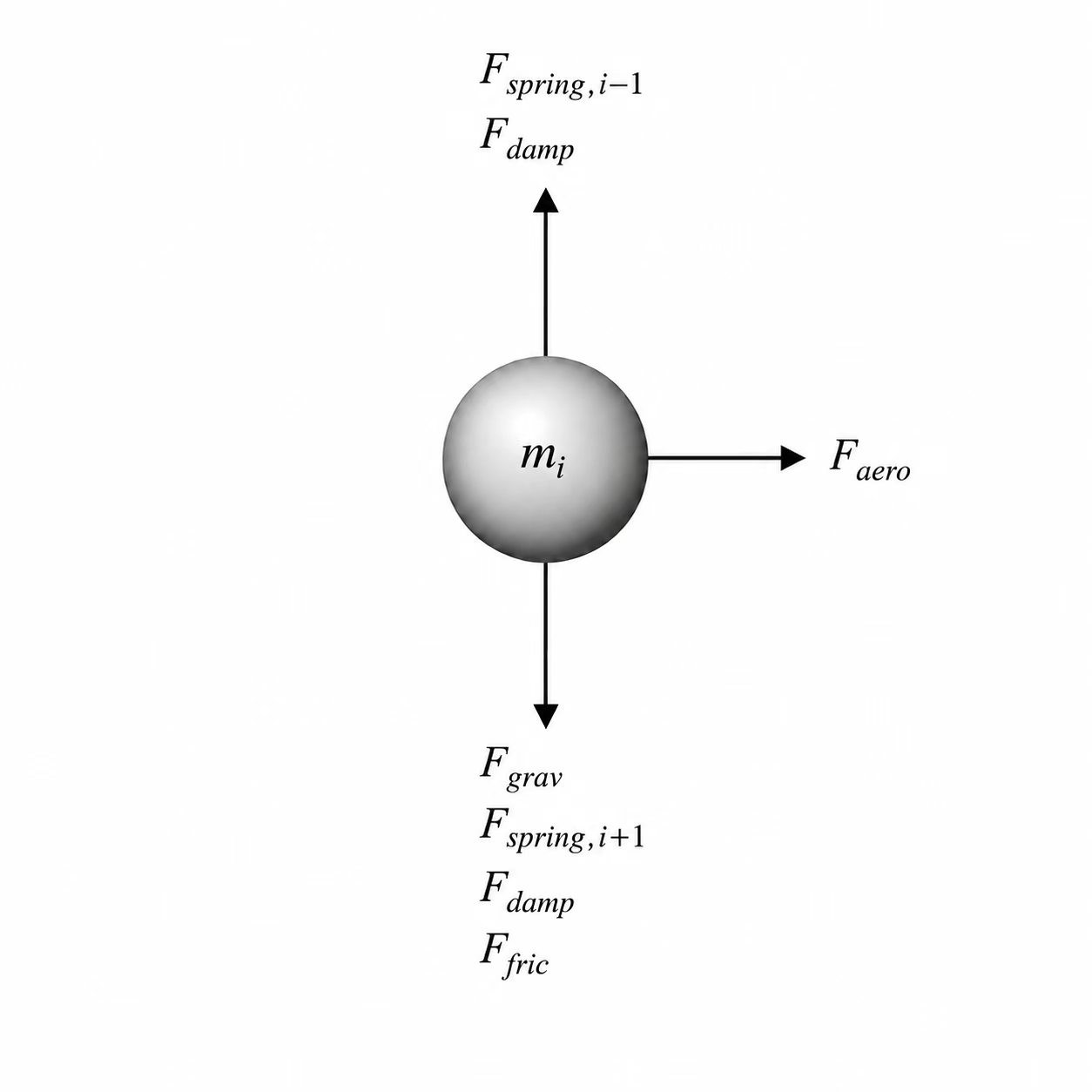}
        \caption{}
        \label{fig:2a}
    \end{subfigure}
    \hfill
    \begin{subfigure}{0.4\textwidth}
        \includegraphics[width=0.9\textwidth, height=2in]{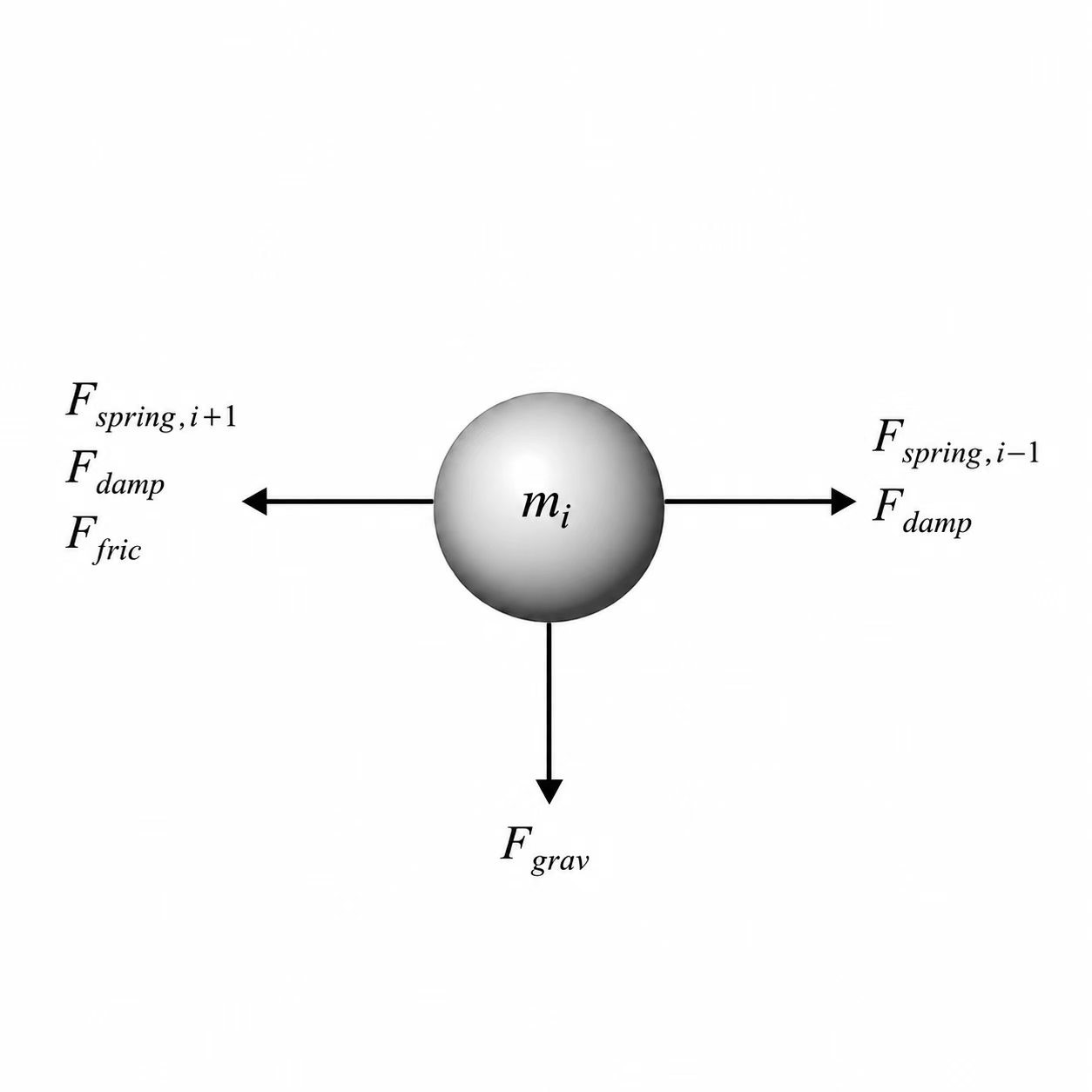}
        \caption{}
        \label{fig:2b}
    \end{subfigure}
    \caption{(\subref{fig:2a}) Single Particle Force Diagram (Upward Pull). (\subref{fig:2b}) Single Particle Force Diagram (Rightward Pull).}
    \label{fig:2}
\end{figure}
For a particle located at arc length s, the resultant force consists of four components, which are described as follows: 
\\(1) Spring force $F_{\text{spring}}$ : When the particle and its adjacent particles are fully deployed with the deployment activation factor $\alpha>0.01$, the spring force acts along the axial direction of the line segment. The resultant spring force on the particle is the difference between the spring forces from the upper and lower adjacent segments. 
\begin{equation}
F_{\text{spring},i}=k\cdot\left(\left|\boldsymbol{r}_{i+1}-\boldsymbol{r}_i\right|-\Delta L_0\right)\cdot\boldsymbol{u}_i-\left(\left|\boldsymbol{r}_i-\boldsymbol{r}_{i-1}\right|-\Delta L_0\right)\cdot\boldsymbol{u}_{i-1}
\label{eq:1}
\end{equation}
where \(u_i\) denotes the unit vector along the direction of the line segment. 
\\(2) Damping force $F_{\text{damp}}$: It is proportional to the component of the relative velocity between adjacent particles along the line segment. 
\begin{equation}
F_{\text{damp},i} = c\cdot\big(\Delta \boldsymbol{v}_i\cdot\boldsymbol{u}_i\big)\cdot\boldsymbol{u}_i 
- \big(\Delta \boldsymbol{v}_{i-1}\cdot\boldsymbol{u}_{i-1}\big)\cdot\boldsymbol{u}_{i-1}
\label{eq:2}
\end{equation}
(3) Gravitational force $F_{\text{grav}}$: Each particle is subjected to a constant gravitational force.
\begin{equation}
F_{\text{grav},i} = m_i \cdot g
\label{eq:3}
\end{equation}
(4) Aerodynamic force $F_{\text{aero}}$: This force only acts on deployed segments and is perpendicular to the line segment. The cross-flow drag model is adopted:
\begin{equation}
F_{\text{grav},i} = m_i \cdot g
\label{eq:4}
\end{equation}
where $v_p$ is the component of particle velocity perpendicular to the line segment. 

The governing equation of particle motion follows Newton's second law. For undeployed particles ($\alpha\approx0$), their positions are constrained by the geometry of the folded path, and neither spring force nor damping force is activated. For particles in the transition zone at the deployment front, the acting forces scale smoothly with $\alpha$.

\subsection{Force Model for Continuous Particle Extraction }

The inflation process of the pilot parachute is described by combining the Pflanz formula and the Gaussian overshoot model \cite{Ref23}. During the suspension line extraction phase, the pilot parachute inflates first, and the inflation duration is calculated via the empirical Pflanz formula. 
\begin{equation}
t_f = \frac{K D_0}{\nu_L^n} = \frac{7 \times 0.66}{55} = 0.084\ \mathrm{s}
\label{eq:5}
\end{equation}
The variation of the product of drag coefficient and reference area \(C_dS\) of the pilot parachute is characterized by a linear rising curve superimposed with a Gaussian pulse. 
\begin{equation}
C_d S(t) = C_d S_{\text{steady}} \cdot \min\big(1,\frac{t}{0.010}\big)
+ \big(C_d S_{\text{peak}} - C_d S_{\text{steady}}\big)
\cdot \exp\left(-\frac12 \left(\frac{t-t_f}{\sigma}\right)^2\right)
\label{eq:6}
\end{equation}
where $C_dS_\text{steady}=0.43\ \mathrm{m}^2$, $C_dS_\text{peak}=6.0\ \mathrm{m}^2$, $t_f=84.0\ \mathrm{ms}$ and $\sigma=25.0\ \mathrm{ms}$. This model generates a drag spike in the form of a Gaussian pulse at the early inflation stage to simulate the impact effect occurring at the instant the pilot parachute opens.

The dynamics of the suspension line extraction process are governed by the variable-mass differential equation. The state variables are the extracted line length \(y(t)\) and extraction velocity \(v_{ext}\), where 
\begin{equation}
v_{\text{ext}} = \frac{dy}{dt}
\label{eq:7}
\end{equation}
\begin{equation}
\frac{dv_{\text{ext}}}{dt}
= -g + \frac{F_{\text{total}}}{m_{\text{drogue,eff}}}
- \frac{F_{\text{fric}}}{m_{\text{drogue,eff}}}
- \frac{\mu \cdot v_{\text{ext}}^2}{m_{\text{drogue,eff}}}
\label{eq:8}
\end{equation}
Here, $m_{\text{drogue,eff}}=M_d+\mu_{\text{total}}\cdot y$ denotes the effective mass on the pilot parachute side. $F_{\text{total}}$ consists of the aerodynamic drag of the pilot parachute, the aerodynamic drag of extracted line segments, and the friction force from the parachute pack. This ordinary differential equation  is integrated from $t=0$ until $y=L$, at which point the line straightening event is triggered.
At the moment of line straightening, the velocity difference $\Delta v$ between the pilot parachute and the payload generates a snatch load via elastic collision:
\begin{equation}
F_{\text{snatch}} = N \cdot \Delta v \cdot D_\zeta \cdot \sqrt{k_\mathrm{single} \cdot M_\mathrm{eff}}
\label{eq:9}
\end{equation}
where $k_{\text{single}}=EA/L$ is the axial stiffness of a single suspension line, $M_{\text{eff}}=M_p+M_dM_pM_d$ is the effective collision mass, and $D_\zeta=\exp\left(-\dfrac{\pi \zeta}{2\sqrt{1-\zeta^2}}\right)$ represents the damping attenuation factor. After straightening, the tension decays over time and finally reaches a steady value, which is determined by the aerodynamic drag of the pilot parachute under the common velocity.

\subsection{Architecture of PINN  }

This model is constructed under a physics-informed neural network (PINN) framework, in which the suspension line is discretized into a mass-point system parameterized by arc-length coordinate $s\in[0,1]$. Taking time $t$ and spatial coordinate $s$ as inputs, the network predicts the deployment state $\alpha(t,s)$, tension $T(t,s)$ and vertical velocity $v_y(t,s)$ of each mass point. Two normalized spatio-temporal variables, $t_\text{norm}=t/T_\text{max}\in[0,1]$ and $s_\text{norm}=s/L\in[0,1]$, are fed into the network, and the three output physical quantities are listed in Table 1.
\begin{table}[H]
\centering
\caption{Output Variables of the PINN Model }
\label{tab:1}
\begin{tabular}{|c|c|c|c|}\hline

Deployment activation factor& $\alpha$ & $\mathbb{R}^+$ (N) &0 for folded particles, 1 for full extraction.\\\hline

Tension of single line& $T$ & $\mathbb{R}^+$ (N) & Tension at different arc-length positions.\\\hline

Velocity in Y direction& $v_y$ & $\mathbb{R}$ (m/s) & Vertical velocity component of the particle\\ \hline

\end{tabular}

\end{table}
\begin{figure}[H]
    \centering
    \includegraphics[width=0.60\linewidth]{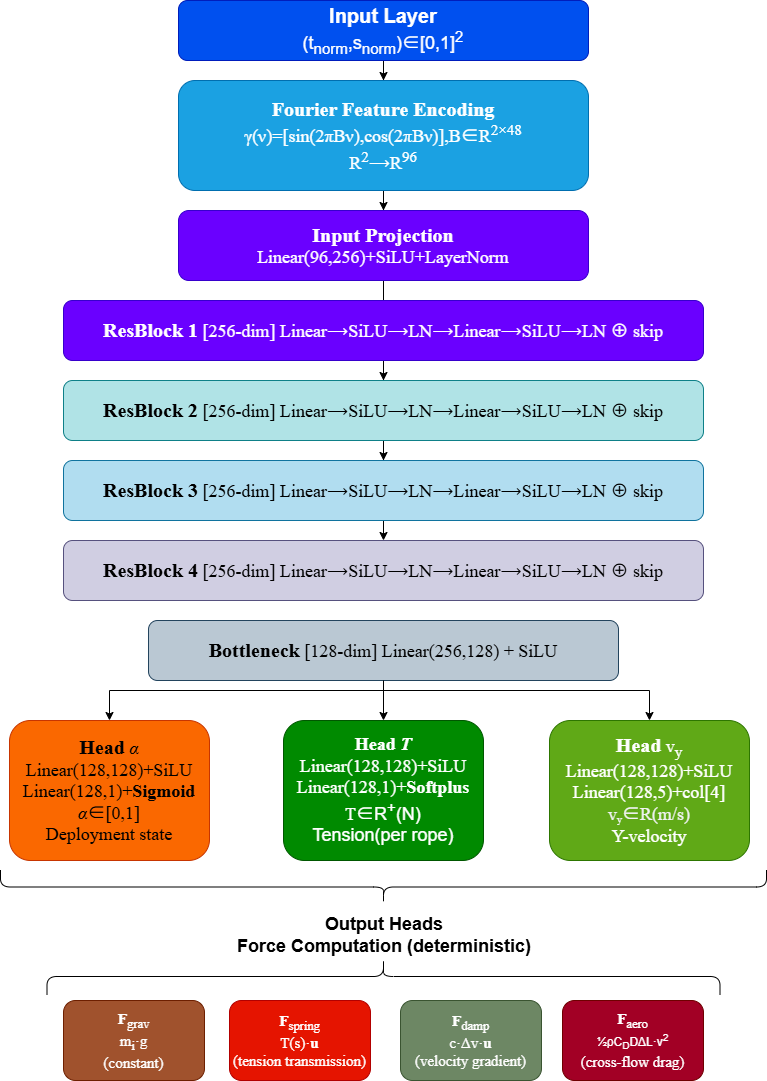}
    \caption{PINN Model Structure Diagram for Suspension Line Extraction Without Binding Tapes }
    \label{fig:3}
\end{figure}
This Physics-Informed Neural Network (PINN) adopts a structure combining Fourier Feature Encoding and Residual Multilayer Perceptron (Residual MLP)\cite{Ref30}. As shown in Figure 3, the network has a total of 323,335 trainable parameters. The detailed architecture is described as follows: 
The Fourier feature encoding layer first maps the two-dimensional temporal-spatial input variables (t,s) into a 96-dimensional feature space via a random Fourier basis. :
\begin{equation}
\gamma(\nu) = \big[\sin(2\pi B\nu),\cos(2\pi B\nu)\big],\quad \nu=(t_\text{norm},s_\text{norm})
\label{eq:10}
\end{equation}

This module effectively extracts high-frequency features from the data and accurately captures short-duration narrowband signals such as the inflation spike of parachute cords. Afterwards, an input projection layer maps the high-dimensional encoded features into a 256-dimensional latent space. 

Four identical residual blocks are stacked in the model. Each block consists of two linear transformations followed by the SiLU activation function and layer normalization. Residual learning is realized via skip connections, which effectively mitigate the gradient vanishing problem in deep networks while retaining the 256-dimensional feature space. A bottleneck layer subsequently compresses the feature dimension to 128 for feature fusion and dimensionality reduction. 

The model adopts a three-branch output architecture that shares the 128-dimensional bottleneck features to predict three key physical quantities separately. The deployment state branch applies the Sigmoid activation to output the deployment coefficient $\alpha$ ranging from 0 to 1. The tension branch utilizes the Softplus activation alongside a scaling factor of 5000 to produce positive parachute cord tension $T$. The velocity branch extracts the corresponding feature channel with a scaling factor of 50 to output the vertical velocity $v_y$.

The network only learns to predict three core variables: $\alpha$, $T$ and $v_y$. All other mechanical quantities are analytically solved by fixed physical formulas without training. Gravity remains a constant value, while spring force, damping force and aerodynamic force are calculated from cord tension, velocity gradient and normal airflow velocity, respectively. This architecture decouples invariant physical constants from spatially and temporally varying quantities to be learned, significantly reducing the learning burden of the network and ensuring physical consistency of predictions.

The training dataset of the model is generated by solving two types of ordinary differential equations (ODEs). The first-stage solution incorporates variable mass, pilot parachute inflation, cord flutter and aerodynamic drag effects, while the second stage is solved based on the dynamic equations of elastic chains. Reference ground truth values of deployment state, tension and velocity are thereby obtained.The total loss function of the model is defined as the weighted sum of normalized mean squared errors (MSEs) corresponding to the three predicted quantities:
\begin{equation}
L_{\text{total}} = 10\cdot \text{MSE}\big(\alpha_{\text{pred}},\alpha_{\text{ref}}\big) 
+ 20\cdot \text{MSE}\left(\frac{T_{\text{pred}}}{5000},\frac{T_{\text{ref}}}{5000}\right) 
+ 2\cdot \text{MSE}\left(\frac{v_{y,\text{pred}}}{50},\frac{v_{y,\text{ref}}}{50}\right)
\label{eq:11}
\end{equation}

To balance the training weights of each physical quantity, the tension and velocity are normalized using the characteristic straightening force of 5000 N and characteristic deployment velocity of 50 m/s, respectively. In line with the physical priority of each output variable, the loss weight for tension is set to the highest value, followed by the weight for deployment state, and the weight for velocity is assigned the minimum value. This weighting scheme prioritizes the prediction accuracy of parachute cord tension and deployment state. 

\subsection{Modelling of Suspension Lines with Binding Tapes  }

Based on the previous research on the PINN model without binding tapes, the constraint mechanism of binding tapes is incorporated into the force model of suspension line particles. The 24.0 m-long suspension line is discretized into a continuous particle system along the arc length. Taking six evenly distributed binding tapes as discrete constraint boundaries, a dynamic model for line extraction is established, in which Substage A (slack extraction with variable-mass ODE) and Substage B (elastic coupling and fracture) proceed alternately. In the PINN framework, binding tape event features are added as explicit inputs. Each binding tape corresponds to a 3-dimensional feature, resulting in a total of 18-dimensional features. An analytical correction strategy based on the \verb|max()| function is adopted during inference to compensate for the smoothing effect of the neural network on millisecond-scale transient responses, as illustrated in Figure 4. 

The suspension line is initially secured inside the parachute pack by binding tapes, and the fracture strength of each binding tape is defined as \(F_{break,strap}\)\cite{Ref31}. The entire extraction process between two adjacent tapes is divided into two substages: 

Substage A (Slack Extraction) : The pilot parachute together with the extracted line segments forms a variable-mass system. The payload is only subjected to gravity, and the suspension line remains slack without tension coupling\cite{Ref32}. The extracted line length $y=x_p-x_d$ increases from the position of the previous binding tape to that of the next one. The governing equation is given as follows: 

\begin{equation}
\frac{dv_d}{dt} = g - \frac{F_{\text{drag}}(t) + F_{\text{rope}}(y) + F_{\text{fric}}(y)}{m(y)} + \frac{\mu \cdot v_{\text{rel}}^2}{m(y)}
\label{eq:12}
\end{equation}
\begin{equation}
\frac{dv_p}{dt} = g
\label{eq:13}
\end{equation}
\begin{equation}
\frac{dy}{dt} = v_p - v_d = v_{\text{rel}}
\label{eq:14}
\end{equation}
\begin{equation}
m(y) = M_{\text{drogue}} + \mu_{\text{total}} \cdot y
\label{eq:15}
\end{equation}
where $F_{\text{drogue}}$ denotes the aerodynamic drag of the pilot parachute, $F_{\text{rope}}$ is the aerodynamic drag of the extracted line segments, and $F_{\text{fric}}$ represents the friction force from the parachute pack. The term $\mu v_{\text{rel}}^2$ stands for the momentum flux term. Substage A terminates when the extracted line length $y$ reaches the position of the next binding tape, and the system then switches to Substage B. 

Substage B (Tensioning to Fracture) : When the extracted suspension line reaches the position of a binding tape, the line is constrained. The pilot parachute and payload are coupled through the elastic suspension line. The elastic tension is expressed as 
\begin{equation}
T = K_i \cdot \max\big(0, y - y_i\big)
\label{eq:16}
\end{equation}
The binding tape fractures once the accumulated tension reaches its fracture strength $F_{\text{break,strap}}$, and the extraction process proceeds to the next section. Here, 
\begin{equation}
K_i = \frac{E \cdot A_{\text{total}}}{y_i}
\label{eq:17}
\end{equation}
refers to the equivalent stiffness of the currently extracted line segments.
The backbone network is cascaded with six residual blocks. Each block consists of a linear layer (512 → 512), a SiLU activation function and Layer Normalization. The total number of trainable parameters is approximately 2.05 million. The SiLU activation function is adopted to guarantee high-order continuity, and Layer Normalization helps stabilize the training of deep networks. The fracture process of binding tapes generally lasts only 4 to 36 ms. As a smooth function approximator, the MLP-based PINN inevitably smooths transient events on the sub-10 millisecond scale. To address this issue, this work proposes an analytical max() correction strategy during inference. Within the loading interval of binding tapes, the lower bound of analytical tension is calculated using known physical parameters, and the final output is determined by taking the maximum value between the PINN prediction and the analytical result.
The formula for calculating the lower bound of analytical tension is presented below: 
\begin{equation}
T_{\text{analytical}}(t,s) = \frac{F_{\text{break}}}{N_{\text{ropes}}} \times l(t) \times d(t) \times a(s)
\label{eq:18}
\end{equation}
where $l(t)\in[0,1]$ is the loading progress, which increases linearly from 0 to 1 over the interval $[t_{\text{reach}},t_{\text{break}}]$. $d(t)$ denotes the exponential decay factor after fracture with a time constant of 1.5 ms. $a(s)=\sigma\left(\frac{s-s_{\text{tape}}}{0.005}\right)$ represents the spatial mask that is activated only above the binding tape. This correction is applied exclusively during inference, without altering the training objectives or the training procedure.

\begin{figure}[!htb]
    \centering
    \includegraphics[width=1.0\linewidth]{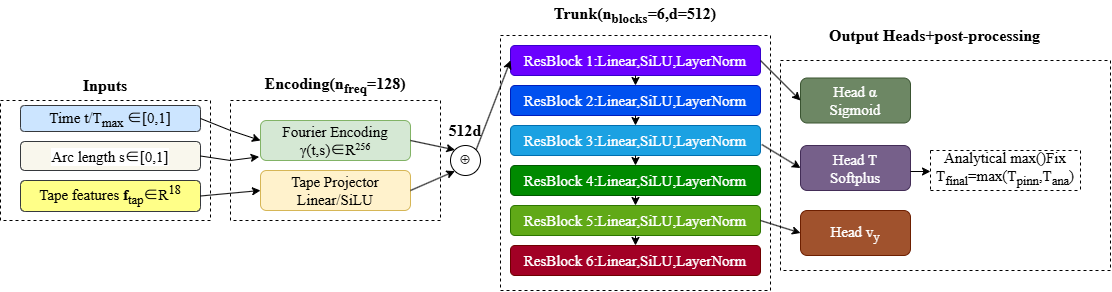}
    \caption{Schematic Diagram of the Model for Suspension Line Extraction with Binding Tapes }
    \label{fig:4}
\end{figure}

\subsection{Model Parameter Settings}

Combined with the practical mechanical properties of parachute cords and relevant research data, all parameters of the model are determined in this paper. Aramid fiber is selected as the material for the parachute cords. The remaining parameters are listed in Table 2.The suspension line is discretized into 122 nodes. The mass of each particle is approximately 0.0001345 kg, the spring stiffness is 2.34 KN/m, and the damping coefficient of the damper is set to 0.05 N·s/m according to the motion characteristics of suspension lines in air\cite{Ref33,Ref34}. As shown in Figure 2, all particles are uniformly distributed inside the parachute pack with an initial velocity of 0 m/s, which is consistent with the actual state before line extraction. 
\begin{table}[H]
    \centering  
    \caption{Main System Parameters }
    \label{tab:2}
\begin{tabular}{cc}  
        \hline
        Parameter & Value \\
        \hline
        Nominal area of pilot parachute / m² & 0.34 \\
        Nominal diameter of main parachute canopy / m & 2.00 \\
        Nominal area of main parachute / m² & 3.14 \\
        Length of single suspension line / m & 24.00 \\
        Number of suspension lines & 10 \\
        Density of suspension line / kg/m³ & 704.80 \\
        Elastic modulus of suspension line / GPa & 1.412 \\
        Air density / kg/m³ & 1.156 \\
        Initial velocity / m/s & 55.00 \\
        \hline
    \end{tabular}
\end{table}

\section{Results And Discussion}

\subsection{Validation of PINN-based Suspension Line Particle Model}

In this study, the trained PINN model is adopted to perform numerical calculations for the extraction process of parachute suspension lines without binding tapes. The obtained results are compared with those from the ODE integration and flight data, as shown in Figure 5. The predicted peak tension time during the pilot parachute inflation is 0.02 s earlier than the experimental value, and the calculated straightening time of suspension lines is 0.01 s shorter than the experimental counterpart. Overall, the numerical results agree well with the experimental data. 
\begin{figure}[H]
    \centering
    \includegraphics[width=0.5\linewidth]{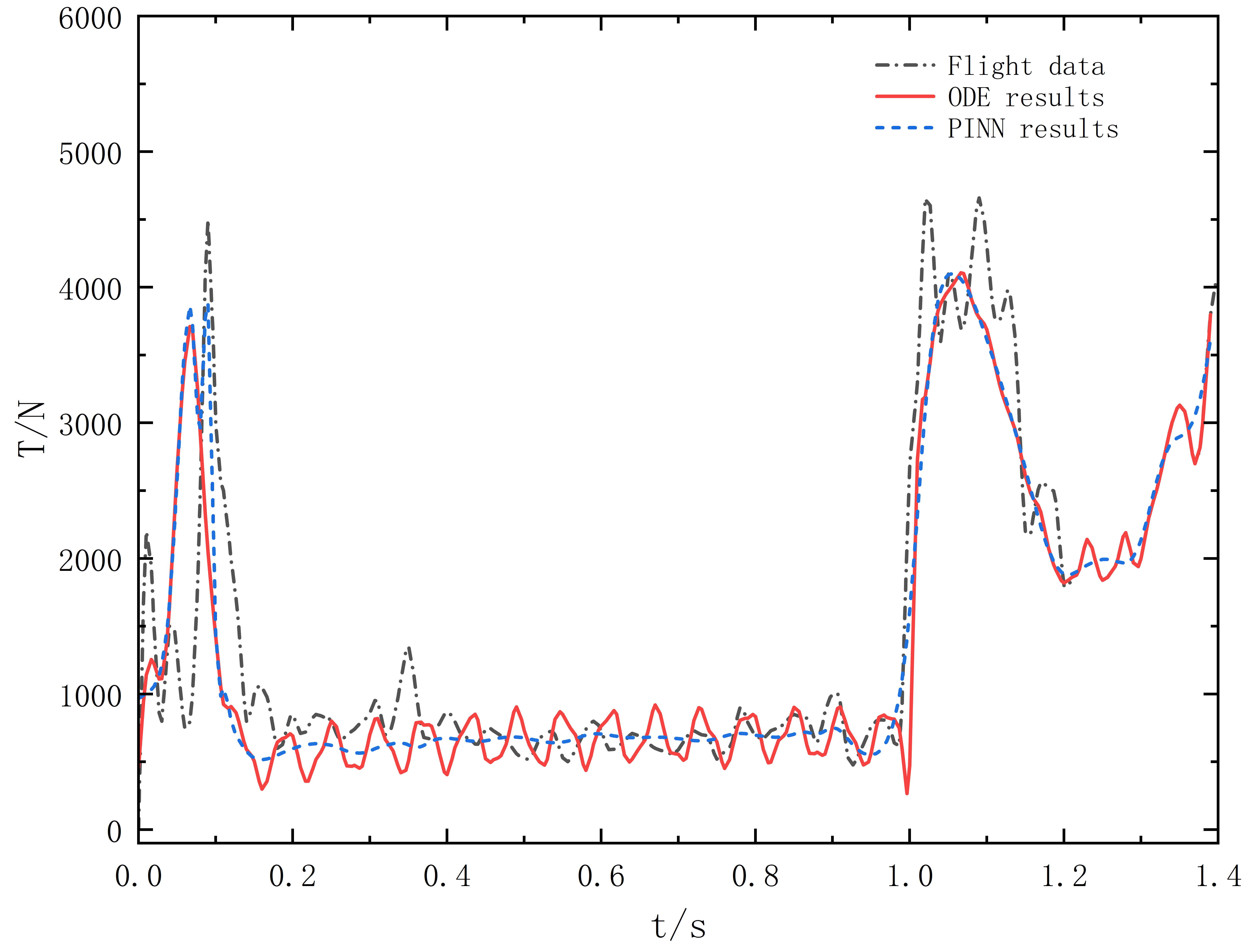}
        \caption{Tension Curve of Suspension Line During Extraction Without Binding Tapes }
    \label{fig:5}
\end{figure}
Ordinary differential equation numerical integration methods such as the Runge-Kutta method are mathematically rigorous and can solve dynamic equations with controllable accuracy\cite{Ref35,Ref36}. However, their computational cost rises sharply with the increase of system degrees of freedom, which limits their application in real-time simulation and parameter sweeping. By embedding physical constraints into neural network training, the well-trained PINN model can predict the state at arbitrary (t,s) points within microseconds, thus achieving remarkable inference efficiency, as presented in Table 3. 

\begin{table}[H]
\centering
\caption{Comparison of Computational Speed Between Two Methods}
\label{tab:3}
\begin{tabular}{cc}
\hline
Method & Single query time \\
\hline
PINN  & $0.77\ \mu\text{s}$\\
ODE   & $154.7s\ \text{s}$\\
\hline
\end{tabular}
\end{table}
\begin{table}[H]
\centering
\caption{Comparison of Calculation Accuracy with the ODE Method }
\label{tab:4}
\begin{tabular}{cc}
\hline
Physical quantity& PINN (\%) \\
\hline
Peak tension during pilot parachute inflation& 0.5 \\
Mean tension at binding tape fracture& $<1.0$\\
Peak snatch load& 0.4 \\
Total extraction time& $<1.0$\\
Velocity difference 
$\Delta v$ & $<1.0$\\
\hline

\end{tabular}

\end{table}
As shown in Table 3, although the total simulation time consumed by the PINN method is longer than that of the ODE method, the PINN method exhibits an overwhelming advantage in the tension inference speed at arbitrary single spatial positions after the full simulation is completed. A single query for the state at arbitrary spatio-temporal coordinates $(t,s)$ only takes $0.77\ \mu\text{s}$, enabling more than 100,000 queries within one second, which satisfies the requirements of real-time simulation and parameter space sweeping.

In contrast, despite the apparent superiority of the ODE method in total computational time, it suffers inherent limitations as it cannot solve the state of individual particles independently. The fast inference capability of PINN stems from the fact that the forward propagation of neural networks only involves matrix multiplication and nonlinear activation operations, without the need for iterative solving or time-stepping integration.Taking the results from the ODE numerical integration method as the benchmark, the relative error is defined as equation 19.
\begin{equation}
\text{Relative Error} = \frac{|\text{PINN result} - \text{ODE reference}|}{\text{ODE reference}} \times 100\%
\label{eq:19}
\end{equation}
 
 Table 4 compares the computational accuracy of the two methods for key physical quantities. The relative errors of the PINN method for all key indicators are kept below 1\%. The finite element method (FEM) achieves accuracy comparable to ODE with sufficiently refined meshes, whereas errors are inevitably introduced when handling contact and fracture behaviors. The mass-spring-damper (MSD) method suffers from relatively low accuracy restricted by the number of particles, though its performance can be improved by increasing particle quantity. 

\subsection{Analysis of the Effect of Binding Tape Quantity on Suspension Line Extraction}

Binding tapes are used to fix the initial configuration of suspension lines and prevent their tangling inside the parachute pack. Their constraints on suspension lines mainly act during the initial state and the extraction process. In this work, dynamic variations in the tension of suspension lines during extraction with and without binding tape constraints are compared and analyzed via numerical simulation, as illustrated in Figure 6.
\begin{figure}[H]
    \centering
    \includegraphics[width=0.5\linewidth]{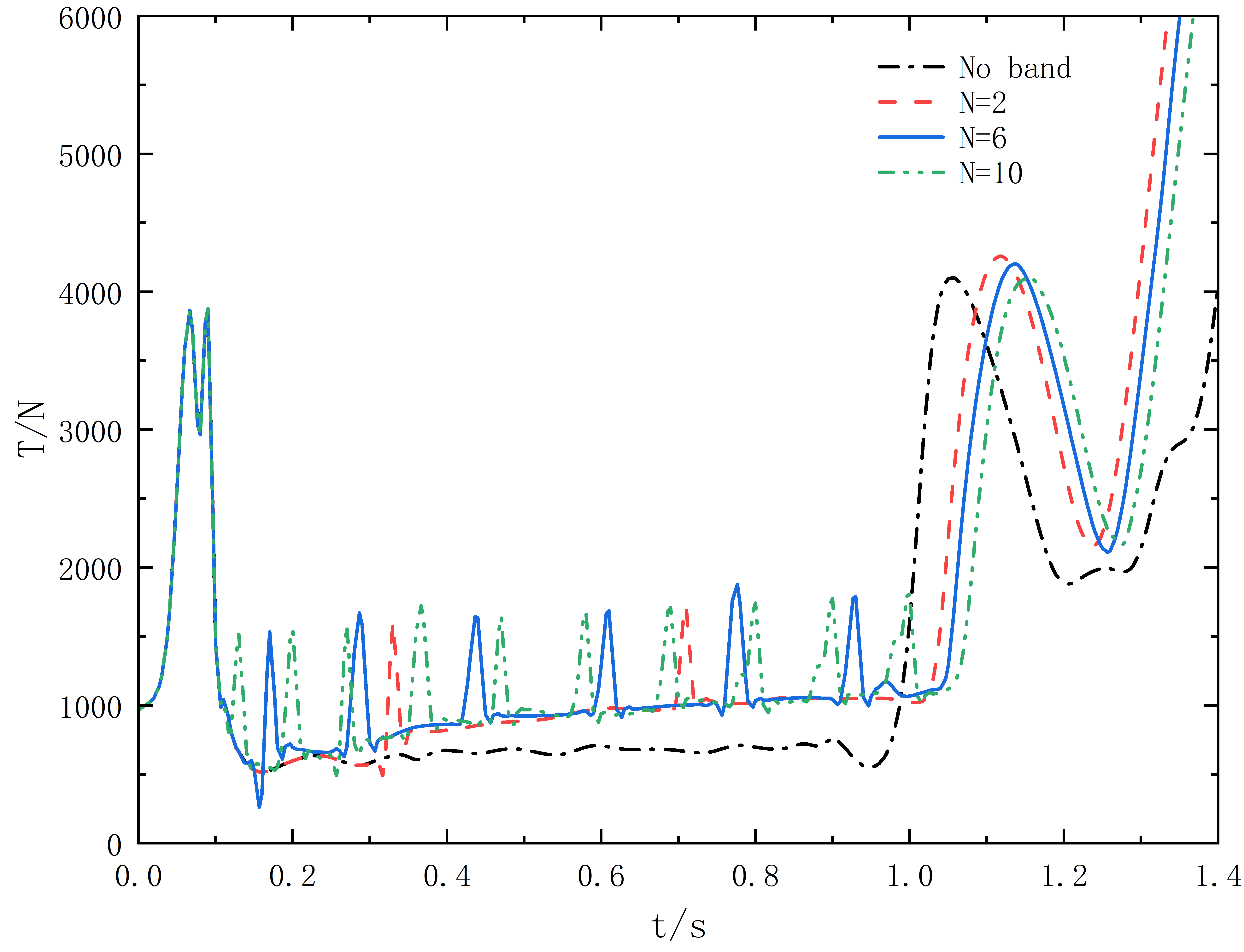 }
    \caption{Tension Comparison for Line Straightening Under Varying Binding Tape Quantities }
    \label{fig:6}
\end{figure}
During the extraction phase, the tension of suspension lines fluctuates more significantly under the constraint of binding tapes. This is because binding tapes introduce additional extraction resistance.The maximum straightening tension is approximately 5.73\% lower than that in the absence of binding tapes. Such results can be attributed to the fact that binding tapes absorb part of the kinetic energy during line extraction, which reduces the velocity difference between the pilot parachute end and the load end.Furthermore, an approximately linear negative correlation exists between the number of binding tapes and the straightening impact. 

Specifically, each additional binding tape reduces the straightening velocity difference $\Delta v$ by about 0.44 m/s and decreases the straightening impact force by around 50 N. This linear relationship performs well when the number of binding tapes $N\le10$, while nonlinear deviation may occur as $N$ further increases. In addition, the straightening time of suspension lines is delayed with the application of binding tapes.

It is also observed that as the number of binding tapes increases, the number of tension pulses during line extraction rises, whereas the pulse amplitude decreases. The average pulse amplitude is 919 N for $N=2$ and 823 N for $N=10$. Meanwhile, the average tension for $N=10$ (1029 N) is higher than that for $N=2$ (891 N). The underlying reason is that a larger number of binding tapes leads to more frequent elastic loading, thereby maintaining a higher average tension level.
\subsection{Analysis of the Effect of Binding Tape Position on Suspension Line Extraction}

The position of binding tapes mainly affects the fracture moment of binding tapes during the extraction of suspension lines. Taking Table 5 as an example, all parameters of the parachute-load system remain identical across five groups of calculations, while only the positions of binding tapes on suspension lines are adjusted to investigate the tension variation during line extraction. Group G1 serves as the control group with evenly spaced binding tapes. Group G2 adopts uniformly arranged binding tapes with a 1 m positional offset on the suspension lines. Groups G3, G4 and G5 correspond to three sets of binding tapes with uneven spacing, where the spacing is restricted within the range of 1 m to 5 m. The specific positional parameters of binding tapes and the calculated straightening forces are presented in Table 5. 

\begin{table}[H]
\centering
\caption{Positions of Binding Tapes and Key Results for Each Group}
\label{tab:5}
\begin{tabular*}{\linewidth}{
p{0.7cm} 
@{\extracolsep{\fill}} 
p{2.2cm} 
p{4.5cm} 
@{\extracolsep{\fill}} 
p{1.8cm} 
p{1.8cm}
}
\hline
\textbf{Group} & \textbf{Distribution} & \textbf{Binding Position (m)} & $\boldsymbol{t_\text{extract}}$~(s) & $\boldsymbol{F_\text{snatch}}$~(N) \\
\hline
G1 & Uniform & 3.4, 6.9, 10.3, 13.7, 17.1, 20.6 & 1.048 & 3860 \\
G2 & Uniform (+1 m offset) & 4.4, 7.9, 11.3, 14.7, 18.1, 21.6 & 1.046 & 3830 \\
G3 & Random 1 & 4.5, 7.9, 11.8, 12.9, 17.8, 22.1 & 1.047 & 3826 \\
G4 & Random 2 & 2.4, 4.9, 8.0, 12.3, 16.1, 19.2 & 1.047 & 3889 \\
G5 & Random 3 & 5.4, 8.2, 10.0, 14.8, 18.3, 20.2 & 1.049 & 3853 \\
\hline
\end{tabular*}
\end{table}
\begin{figure}[H]
    \centering
    \includegraphics[width=0.5\linewidth]{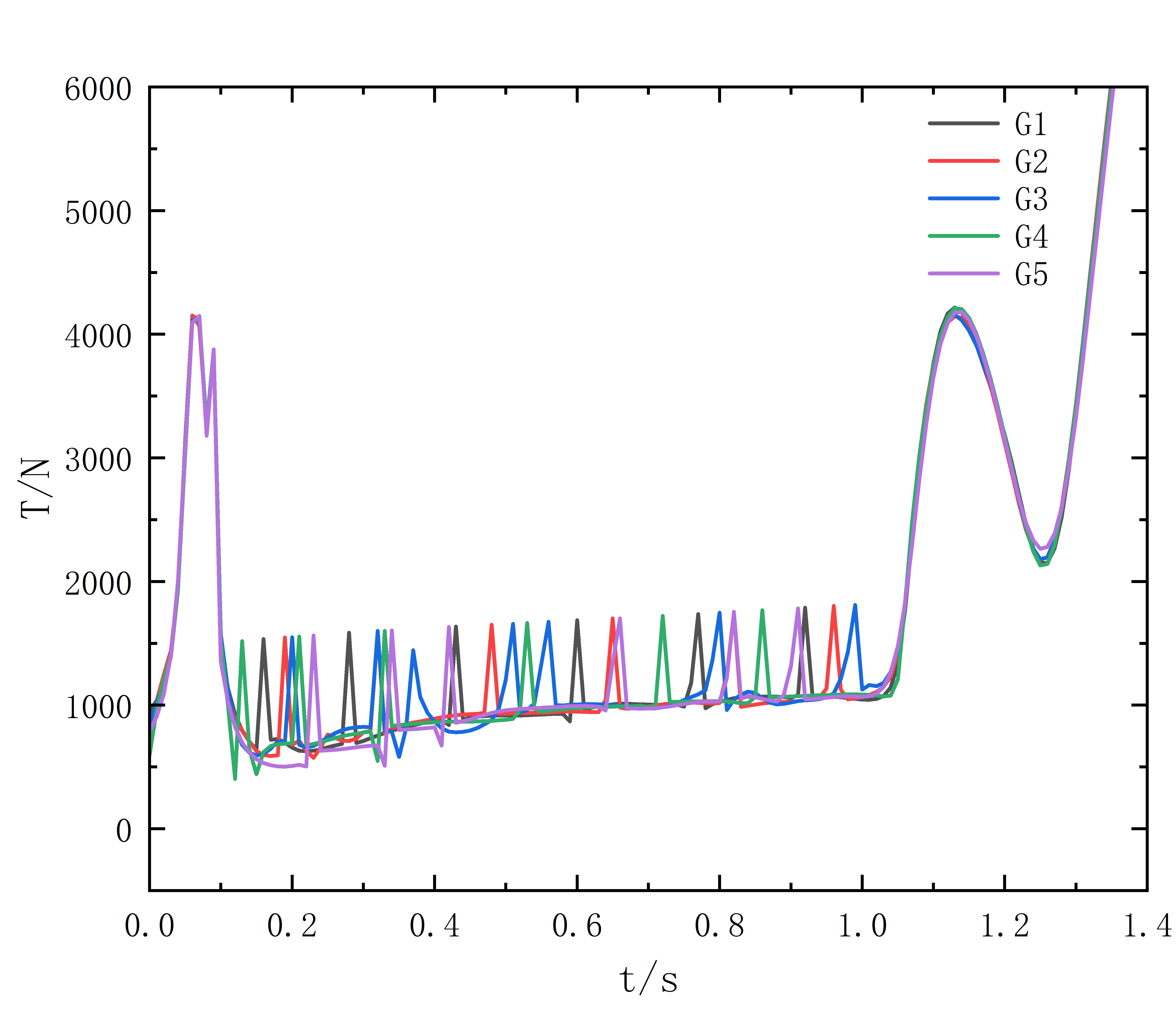}
    \caption{Calculated Tension Results of Suspension Lines under Five Binding Tape Positions }
    \label{fig:7}
\end{figure}
As shown in Figure 7, the distribution forms of binding tapes (uniform, offset and random) have negligible effects on the amplitude of tension pulses generated by single fracture. The average pulse amplitude of the five groups ranges from 742~N to 804~N, with a range of only 62~N. For the  group G3, the spacing between the third and fourth binding tapes is merely 1.1~m, so two fractures occur almost consecutively. The subsequent loading takes place before the previous tension fully dissipates, resulting in abnormal local tension superposition. 

In addition, the straightening forces of suspension lines are nearly identical across all five groups.It can be concluded from the five sets of calculations that the pulse amplitude does not change significantly with the tape positions. This indicates that the fracture threshold $F_\text{tape-break}=1500~\text{N}$ acts as the dominant factor for tape failure. The position of binding tapes mainly affects the fracture time and has a minor influence on the tension increment of a single fracture. Accordingly, it is recommended that the minimum spacing between adjacent binding tapes should be no less than 2~m in engineering design.

\subsection{Effect of Binding Tape Breaking Strength on Suspension Line Extraction}

The mainstream materials for binding tapes used in current parachutes include Nylon 66, ultra-high molecular weight polyethylene and aramid. These materials possess high strength and excellent toughness. Generally, binding tapes are attached to the rings of the parachute pack to restrain suspension lines inside the pack. When the suspension lines are extracted, the binding tapes are subjected to tensile loads and will fracture once the applied force reaches their breaking strength. Therefore, investigating the influence of binding tape strength on the extraction process of suspension lines is of great necessity.

\begin{figure}[H]
    \centering
    \includegraphics[width=0.5\linewidth]{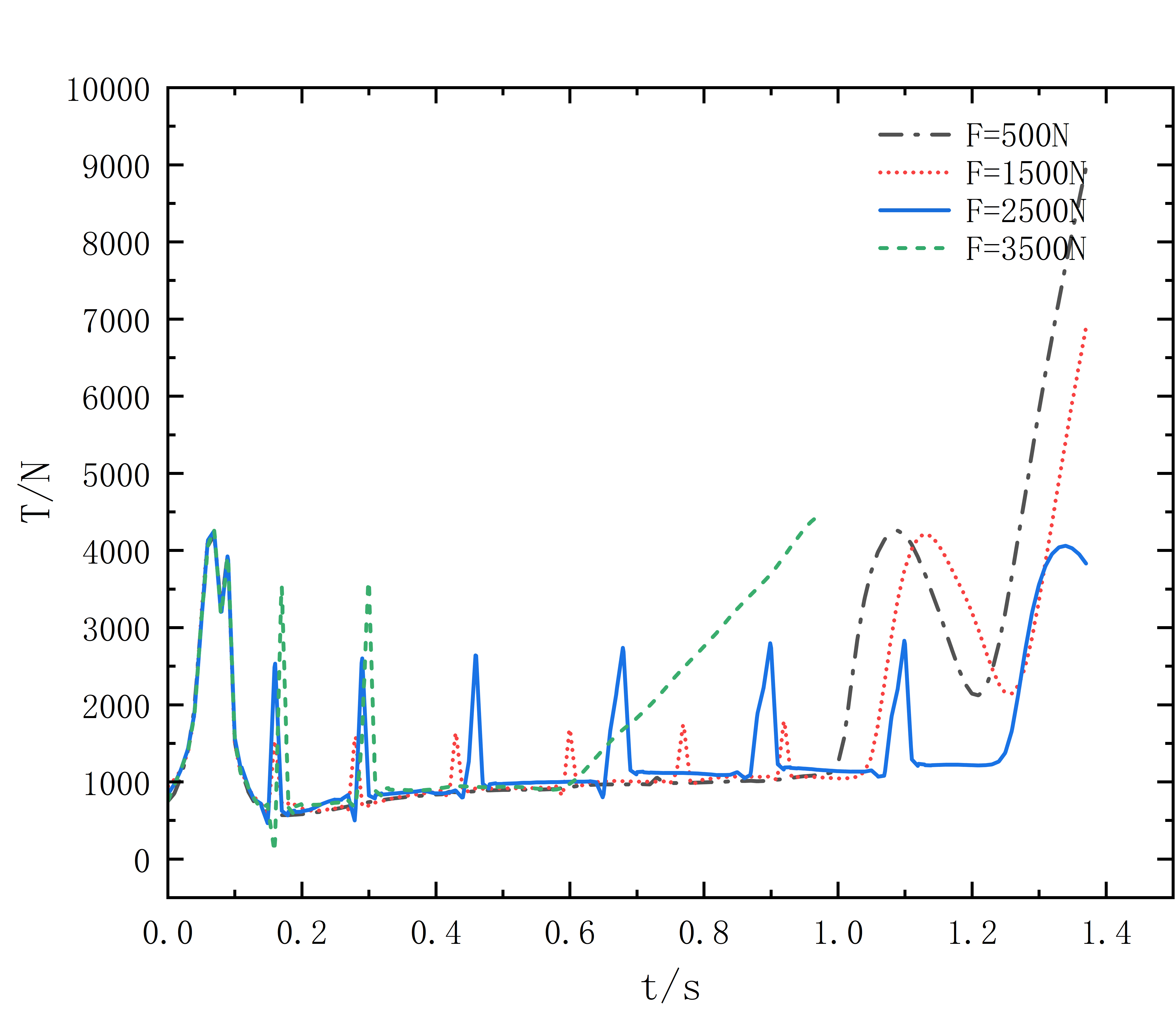 }
    \caption{Calculated Tension Results of Suspension Lines under Four Different Binding Tape Breaking Forces }
    \label{fig:8}
\end{figure}
\begin{table}[H]
\centering
\caption{Table Comparison of Binding Tape Breaking Strength }
\label{tab:6}
\begin{tabular}{ccccc}
\hline
\textbf{Index}& \textbf{F=500N} & \textbf{F=1500N} & \textbf{F=2500N} & \textbf{F=3500N} \\
\hline
Extraction time& 1.000s & 1.048s & 1.249s & 0.981s(Stagnation)\\
$\Delta v_\text{straightening}$ & 26.87 m/s & 25.60 m/s & 22.12 m/s & 17.01 m/s \\
Straightening tension& 15128N & 14458N & 8389N & 4570N \\
Fracture status& 6/6 & 6/6 & 6/6 & 2/6(Stagnation)\\
Phase B of Tape 6& 0.82ms & 13.27ms & 37.03ms & —(Not reached)\\
\hline

\end{tabular}

\end{table}
As shown in Figure 8, the tension responses of the suspension line deployment process under four groups of binding tape strengths are calculated separately. Discrepancies can be observed in the first tension pulse across all curves. The fourth group features a higher breaking force of binding tapes, which dissipates more kinetic energy. Consequently, the fracture time of its binding tapes is slightly delayed relative to the other three groups, accompanied by a larger amplitude of tension pulse. This phenomenon slightly reduces the velocity difference between the pilot chute and payload, leading to insufficient driving force for subsequent suspension line extraction and a delayed fracture instant of the second binding tape. 

Meanwhile, the high-strength binding tape absorbs substantial kinetic energy of the pilot chute, resulting in incomplete extraction of suspension lines and an earlier onset of main parachute inflation. As listed in Table 6, the suspension lines under the condition of  $F=500~\text{N}$ achieve full straightening earliest, with inconspicuous tension fluctuations when the binding tapes rupture. This is mainly attributed to the average force of 866 N acting on each particle during suspension line extraction, which exceeds the breaking strength of the binding tape. 

In addition, the earliest straightening yields a larger velocity difference between the pilot chute and payload at the straightening instant, and the corresponding suspension line tension at this moment is 1.6\% and 4.59\% higher than those of the other test groups. The above results demonstrate that there exists a critical breaking strength $F_\text{crit}$, which is correlated with the stiffness of line segments and the drag force of the pilot parachute. For the current system parameters, $F_\text{crit}$ is approximately in the range of $2500~\text{N}$ to $3000~\text{N}$. When the breaking strength exceeds this critical value, partial binding tapes fail to fracture, resulting in an incomplete extraction process. In engineering design, the breaking force $F_\text{break}$ is recommended to be set within $50\%$ to $80\%$ of $F_\text{crit}$ (i.e., $1250~\text{N}$–$2000~\text{N}$). This configuration ensures that all binding tapes can fracture normally while providing a sufficient deceleration effect.
\subsection{Force Analysis of Particles at Different Positions on Suspension Lines}

The particle model of suspension lines guided by the PINN established in this study can calculate the force acting on any particle of the suspension line at any moment. In this subsection, the particle forces of suspension lines during extraction are calculated for two cases: with and without binding tape constraints.

As shown in Figure 9a, the time-dependent force of particles on the suspension lines is computed at a normalized distance interval of 0.1. It can be observed that the particles near the pilot parachute end are first subjected to an increasing force transmitted by the inflated pilot parachute, which then sequentially pulls out the subsequent particles. Since the suspension lines are placed in a bent configuration inside the parachute pack in the initial state, the initial tension of line particles is non-zero. For the Z-shaped arranged suspension lines, the bending regions exhibit the maximum force change rate, where the tension rises to 500 N–700 N within approximately 30 ms.In the early extraction stage, the pilot parachute is in the dragging phase accompanied by inflation pulses, generating a large drag force and resulting in a fast extraction speed. During the middle and late extraction stages (0.6$<$t$<$1), the drag force of the pilot parachute stabilizes. 

Meanwhile, the aerodynamic resistance and friction of the extracted line segments increase continuously as the lines are pulled out, leading to a gradual reduction in extraction velocity. At the instant when the entire suspension line is fully straightened, the overall tension surges sharply from around 500 N at the late extraction stage to 4095 N. The tension difference between the load end and the pilot parachute end is less than 5\%, indicating nearly uniform synchronous loading along the whole line.
\begin{figure}[H]
    \centering
    \begin{subfigure}{0.45\textwidth}
        \includegraphics[width=\textwidth]{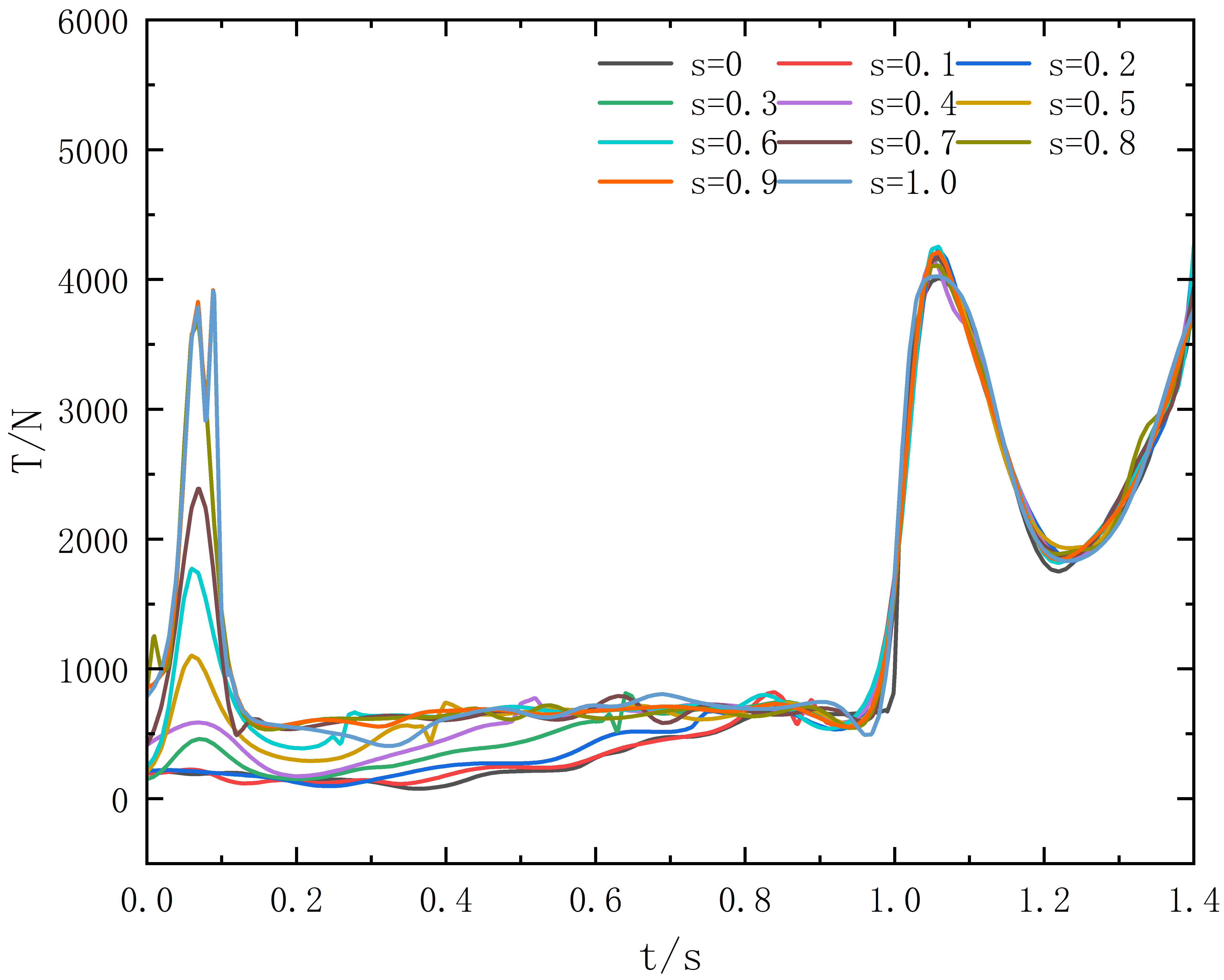}
        \caption{}
        \label{fig:9a}
    \end{subfigure}
    \hfill
    \begin{subfigure}{0.45\textwidth}
        \includegraphics[width=\textwidth]{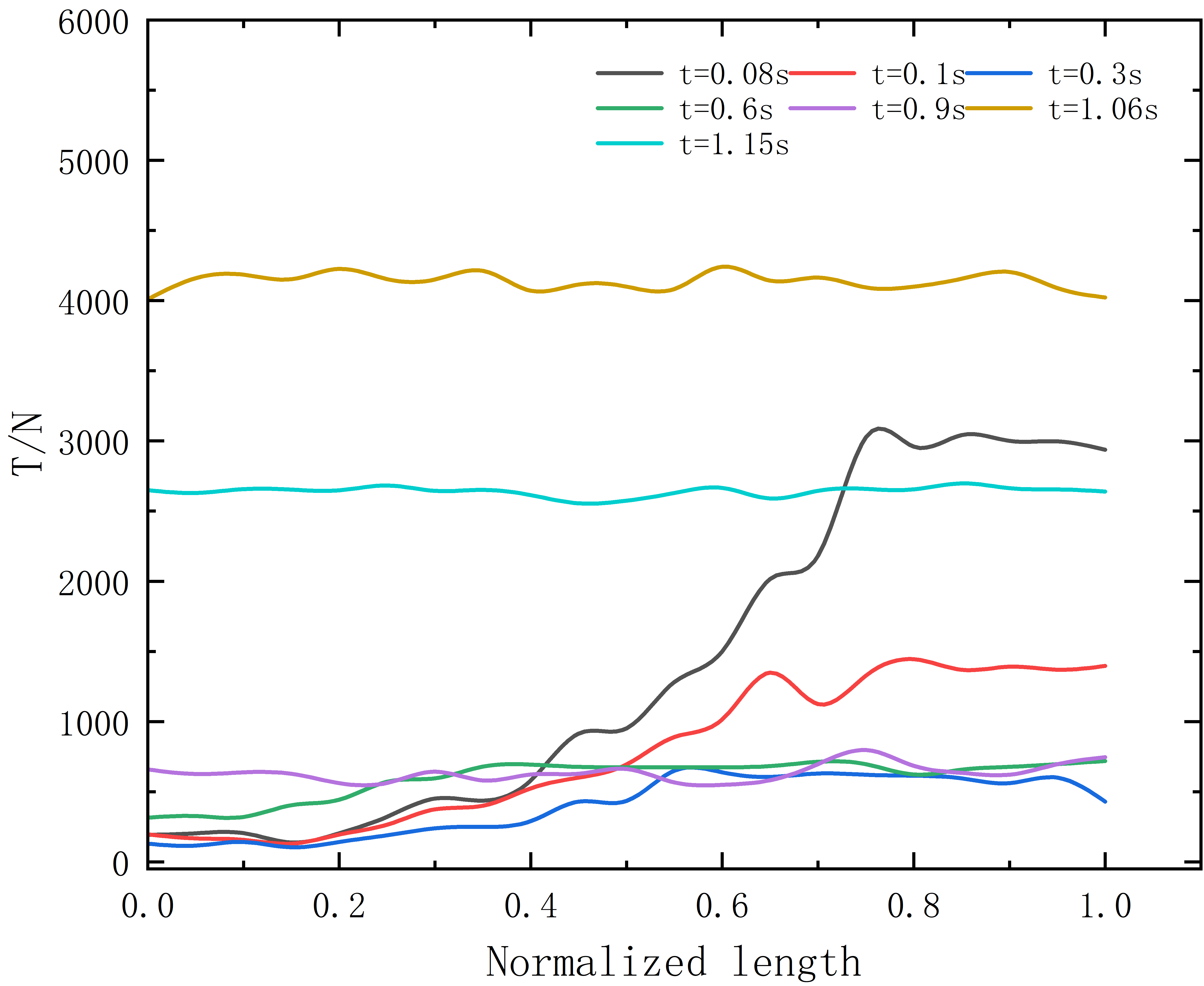 }
        \caption{}
        \label{fig:9b}
    \end{subfigure}
    \caption{(\subref{fig:9a}) Tension variation of particles on suspension lines over time. (\subref{fig:9b}) Force distribution of suspension line particles at different moments.}
    \label{fig:9}
\end{figure}
Figure 9b presents the force distribution of line particles at typical moments, which intuitively illustrates the propagation of forces from the pilot parachute along the suspension lines. The suspension lines are fully straightened at approximately 1.0 s, and the peak straightening tension occurs at 1.06 s. At this moment, the force acting on most particles of the suspension lines exceeds 4 kN.
\begin{figure}[H]
    \centering
    \begin{subfigure}{0.45\textwidth}
        \includegraphics[width=\textwidth]{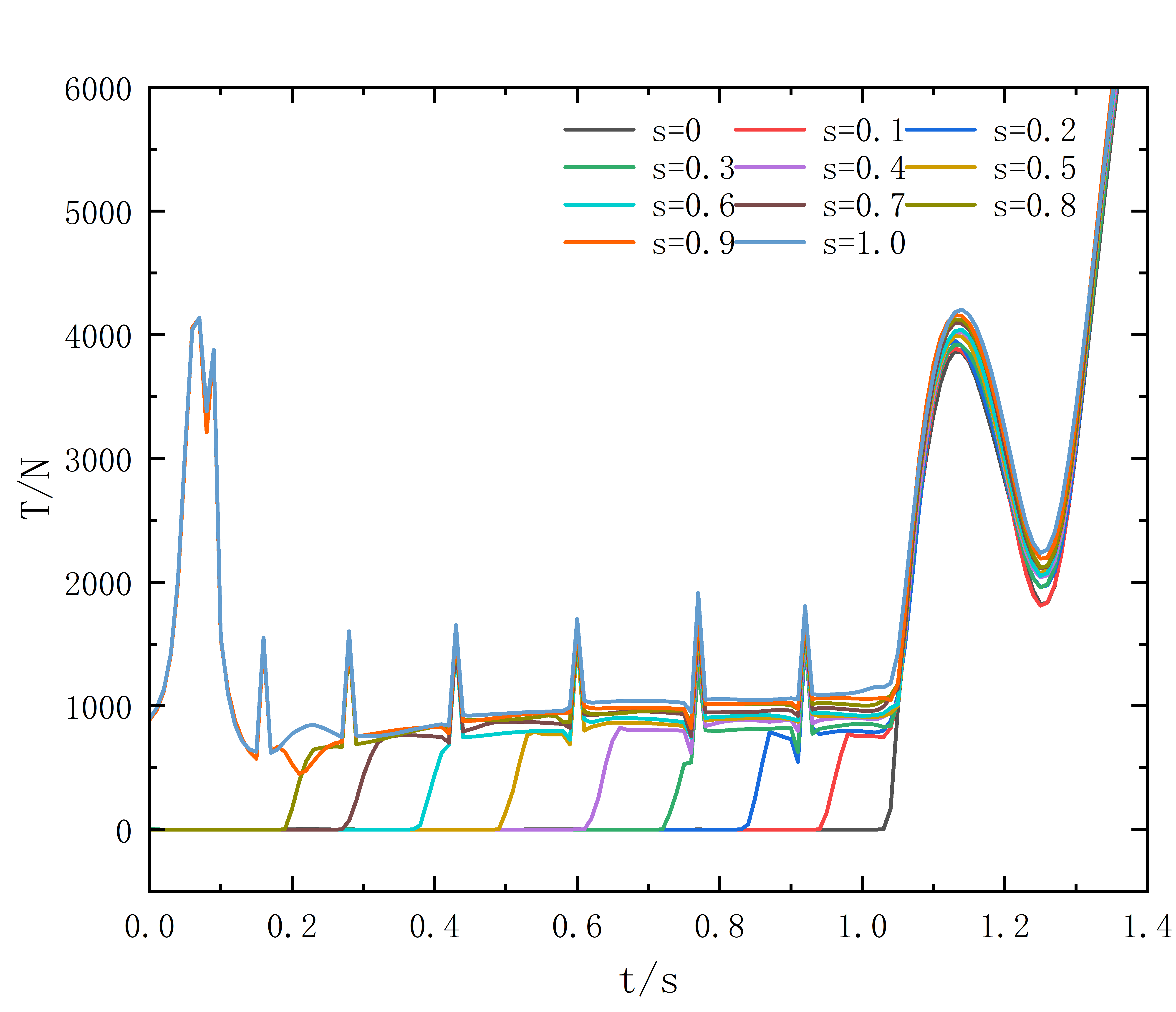}
        \caption{}
        \label{fig:10a}
    \end{subfigure}
    \hfill
    \begin{subfigure}{0.45\textwidth}
        \includegraphics[width=\textwidth]{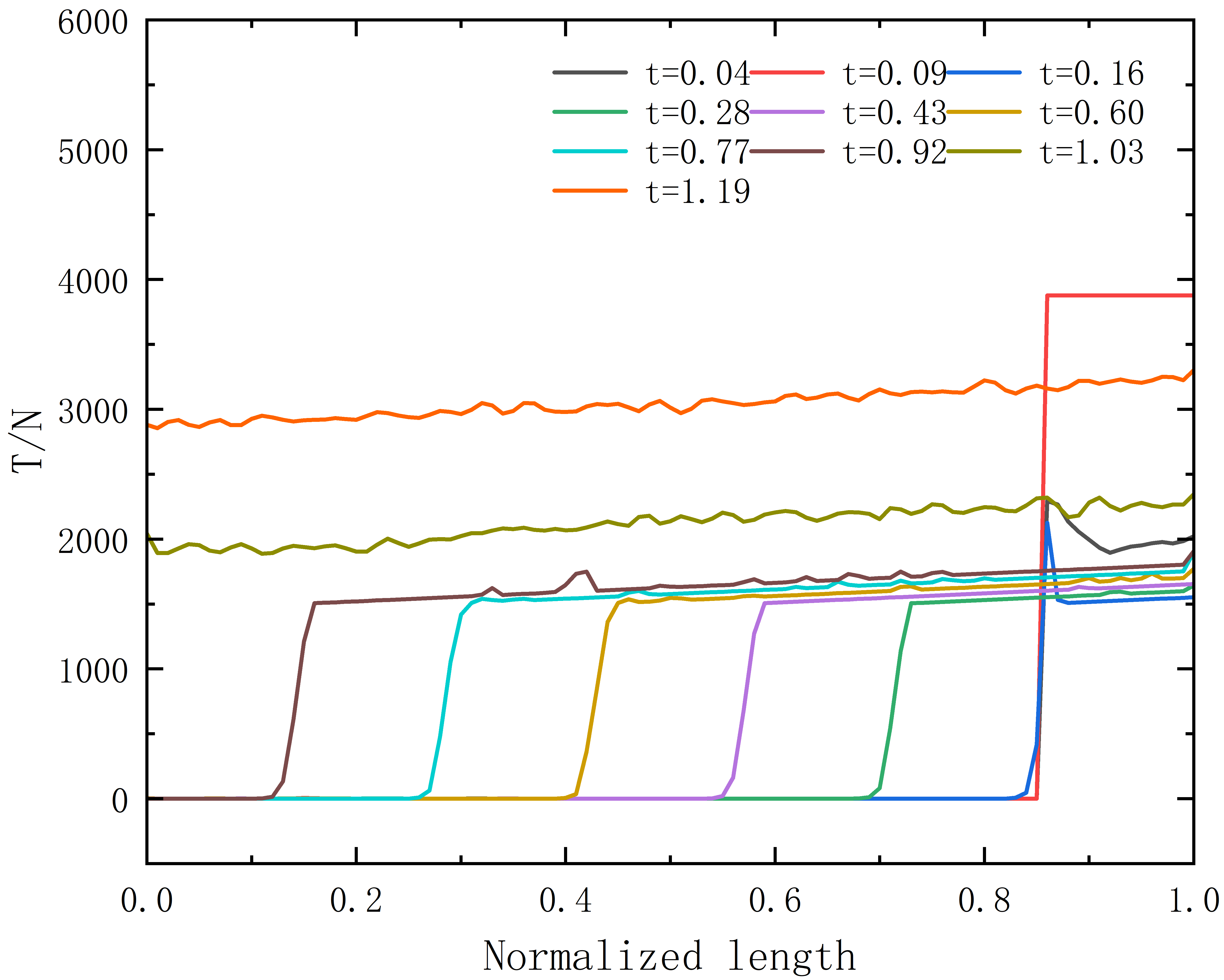}
        \caption{}
        \label{fig:10b}
    \end{subfigure}
    \caption{(\subref{fig:10a}) Tension variation of particles on suspension lines with binding tapes over time. (\subref{fig:10b}) Force distribution of suspension line particles at binding tapes at different moments.}
    \label{fig:10}
\end{figure}
For suspension lines fitted with binding tapes, the tapes hinder tension propagation. It can be clearly observed that the line segments restrained by binding tapes remain unloaded. When a binding tape is about to fracture, the adjacent line segment being extracted starts to bear tension, and the force acting on particles decreases with increasing distance from the tape. This is mainly because the damping between line particles slows down force transmission.After extraction, the tension on particles rises transiently when passing through binding tapes. Such force characteristics are consistent with the Stage A and Stage B defined in Section Modelling of Suspension Lines with Binding Tapes, as illustrated in Figure 10a.Figure 10b shows the calculated force results of particles near binding tapes at different moments. A distinct peak appears on the curve at t=0.16s, which corresponds to the position of the nearest binding tape. The peak force generated by pilot parachute inflation propagates to this particle. Restrained by the binding tape, the tension continues to rise until the tape fractures. 
\section*{Conclusion}
This paper proposes a mechanical modeling method for suspension line deployment based on supervised learning and Physics-Informed Neural Networks (PINN). In this method, the force state of a single particle is established. Fourier feature encoding and residual multi-layer perceptron (MLP) networks are adopted to learn the mapping from spatio-temporal coordinates to mechanical states, and the established model is further extended to the entire suspension line. This study investigates the extraction process of parachute suspension lines. By comparing the effects of the number, position and breaking strength of binding tapes on the tension during line extraction, and calculating and analyzing the force of particles on suspension lines at different moments, the main conclusions are drawn as follows.

(1) The presented model achieves precise predictions of the deployment status, tension and velocity of suspension lines at arbitrary spatial and temporal points. It exhibits favorable prediction accuracy at critical mechanical feature positions throughout the deployment process, with minor deviations in tension and velocity outputs compared with reference results.

(2) Increasing the number of binding tapes can effectively reduce the discrepancy in suspension line straightening velocity and alleviate the transient impact load arising during the straightening process, while inducing more tension fluctuations with diminished peak magnitudes. Nevertheless, an excessive quantity of binding tapes or excessively high tensile strength of the tapes will substantially raise the extraction resistance of suspension lines, thereby impeding their full deployment.

(3) The tension change rate of line particles is remarkably greater under working conditions equipped with binding tapes compared with the tape-free case, which imposes stricter demands on the dynamic mechanical performance of suspension line materials.

This study presents a tension calculation method based on supervised learning-based PINN for simulating the extraction process of parachute suspension lines from the pack.Compared with traditional ODE-based approaches, the proposed method achieves higher computational efficiency and better accuracy. It enables real-time force calculation for each particle on the suspension line and is applicable to different types of main parachutes.The results verify that binding tapes can effectively reduce the peak tension during line straightening. This paper also investigates the influences of the number, layout and breaking of binding tapes on the line extraction process. The developed method provides a useful reference for parachute design and offers insights for research on general cable-related problems.Nevertheless, the complete flight sequence from suspension line extraction to system landing is highly complicated. The present work only investigates the stage covering suspension line extraction and straightening up to the instant before main parachute inflation. Future research can be extended to the full tension calculation of suspension lines throughout the entire deceleration and descent process. In addition, crosswind parameters should be incorporated into the model to improve the consistency between numerical predictions and practical operating conditions. 
\section*{Acknowledgments}

\subsection*{Author Contributions} 

Ronghui Quan proposed and conducted the research; Xiang Zhao established the simulation platform and completed all numerical calculations; Yaqi Xiao and Junlin Chen participated in the simulation calculations. 
All authors contributed equally to the writing of the manuscript.

\subsection*{Funding}
This work was supported by the National Natural Science Foundation of China (No. 42241148) and the Open Foundation of Laboratory of Aerospace Entry, Deceleration and Landing Technology . 

\subsection*{Conflicts of Interest}
The authors declare that there is no conflict of interest regarding the publication of this article.

\subsection*{Data Availability}
The data include the force calculation codes for suspension line particles, parameter sets of the PINN model, and numerical simulation results. All datasets generated and analyzed throughout this study are available from the corresponding author upon reasonable request. 

\section*{Supplementary Materials}
The authors declare that no supplementary materials accompany this article. A complete list of group authors is provided below. 

\section*{Guidelines for References}

\bibliographystyle{unsrtnat}
\bibliography{sample}

\begin{thebibliography}{39}
\providecommand{\natexlab}[1]{#1}
\providecommand{\url}[1]{\texttt{#1}}
\expandafter\ifx\csname urlstyle\endcsname\relax
  \providecommand{\doi}[1]{doi: #1}\else
  \providecommand{\doi}{doi: \begingroup \urlstyle{rm}\Url}\fi

\bibitem[Zhao et~al.(2025)Zhao, Zhou, Zhao, et~al.]{Ref1}
C.~Zhao, Y.~Zhou, S.~Zhao, et~al.
\newblock Research progress on the dynamics of parachute deceleration systems.
\newblock 3109\penalty0 (1):\penalty0 012028, 2025.

\bibitem[Pantano and Montanez(2024)]{Ref2}
C.~Pantano and R.~Montanez.
\newblock Fluid-structure inflation simulations of a parachute with high-speed dynamic reefing.
\newblock 2024.

\bibitem[Tang et~al.(2021)Tang, Wang, Liu, et~al.]{Ref3}
M.~Tang, L.~Wang, Y.~Liu, et~al.
\newblock Effects of aerodynamics on line sail during parachute deployment.
\newblock 2021.

\bibitem[Lu(2014)]{Ref4}
Y.~Lu.
\newblock Dynamic modeling of parachute deployment in mars environment.
\newblock 2014.

\bibitem[Sadeck and Lee(2009)]{Ref5}
J.~Sadeck and C.~K. Lee.
\newblock Continuous disreefing method for parachute opening.
\newblock 46:\penalty0 501--504, 2009.

\bibitem[Bergeron et~al.(2024)Bergeron, Ghoreyshi, and Jirásek]{Ref6}
K.~Bergeron, M.~Ghoreyshi, and A.~J. Jirásek.
\newblock Simulation and stability analysis of a coupled parachute-payload system.
\newblock 2024.

\bibitem[Xing et~al.(2020)Xing, Li, Chen, et~al.]{Ref7}
X.~Xing, F.~Li, X.~Chen, et~al.
\newblock Research on the inflation process of the drag parachute in the landing of first-stage booster and its key parameters.
\newblock 97:\penalty0 5057--5062, 2020.

\bibitem[Yu et~al.(2022)Yu, Wang, and Pantano]{Ref8}
H.~Yu, Q.~Wang, and C.~Pantano.
\newblock Inflation simulations of a conical ribbon parachute at subsonic conditions.
\newblock 2022.

\bibitem[Jolly(1966)]{Ref9}
A.~G. Jolly.
\newblock A method of investigating the deployment characteristics of man-carrying parachutes.
\newblock Defense Technical Information Center (DTIC), 1966.

\bibitem[Brandeau et~al.(2017)Brandeau, Sanchez, Owens, et~al.]{Ref10}
E.~Brandeau, J.~G. Sanchez, A.~Owens, et~al.
\newblock Device for temporal measurement of loads in parachute suspension systems.
\newblock NASA Technical Reports Server (NASA), 2017.

\bibitem[McVey and Wolf(1974{\natexlab{a}})]{Ref11}
D.~F. McVey and D.~F. Wolf.
\newblock Analysis of deployment and inflation of large ribbon parachute.
\newblock 11, 1974{\natexlab{a}}.

\bibitem[Sundberg(1993)]{Ref12}
W.~D. Sundberg.
\newblock Status report: Parachute system design, analysis and simulation tool.
\newblock AIAA Paper 93-1208, 1993.

\bibitem[Moog(1975)]{Ref13}
P.~D. Moog.
\newblock Aerodynamic line blowing parachute deployment.
\newblock AIAA Paper 75-1381, 1975.

\bibitem[Srinivas(1995)]{Ref14}
E.~K. Srinivas.
\newblock Modeling issues related to retrieval flexible tethered satellite systems.
\newblock 18:\penalty0 1169--1176, 1995.

\bibitem[Moog(1973)]{Ref15}
R.~D. Moog.
\newblock Qualification flight tests of the viking decelerator system.
\newblock AIAA Paper 73-457, 1973.

\bibitem[Peng et~al.(2014)Peng, He, and Wang]{Ref16}
W.~Peng, Y.~He, and T.~Wang.
\newblock Granular temperature with discrete element method simulation in a bubbling fluidized bed.
\newblock 25:\penalty0 896--903, 2014.

\bibitem[Navarro and Braun(2013)]{Ref17}
H.~A. Navarro and M.~P.~S. Braun.
\newblock Determination of the normal spring stiffness coefficient in the linear spring–dashpot contact model of discrete element method.
\newblock 246:\penalty0 707--722, 2013.

\bibitem[Tiwari et~al.(2025)Tiwari, Bose, and Kumaran]{Ref18}
A.~K. Tiwari, M.~Bose, and V.~Kumaran.
\newblock Energy non-equipartition in vibrofluidized particles.
\newblock 2025.

\bibitem[Führer et~al.(2024)Führer, Brendel, and Wolf]{Ref19}
F.~Führer, L.~Brendel, and D.~E. Wolf.
\newblock Correction of the spring-dashpot-slider model.
\newblock 26, 2024.

\bibitem[Xu(2002)]{Ref37}
Zhenlong Xu.
\newblock Parallel finite element methods for modeling contact in geometrically nonlinear membrane structures.
\newblock Dissertation, 2002.

\bibitem[Stein(1999)]{Ref38}
K.~R. Stein.
\newblock Simulation and modeling techniques for parachute fluid-structure interactions.
\newblock Dissertation, 1999.

\bibitem[Takizawa and Tezduyar(2012)]{Ref39}
Kenji Takizawa and Tayfun~E. Tezduyar.
\newblock Computational methods for parachute fluid-structure interactions.
\newblock 19:\penalty0 125--169, 2012.

\bibitem[Mowlavi and Nabi(2022)]{Ref40}
S.~Mowlavi and S.~Nabi.
\newblock Optimal control of pdes using physics-informed neural networks.
\newblock 473:\penalty0 111731, 2022.

\bibitem[Watts(1993)]{Ref20}
G.~Watts.
\newblock Space shuttle solid rocket booster main parachute damage reducing team report.
\newblock NASA TM-4437, 1993.

\bibitem[French(1984)]{Ref21}
K.~E. French.
\newblock Parachute canopy dimpling collapse mode.
\newblock AIAA Paper IAA-84-0796, 1984.

\bibitem[Toni(1970)]{Ref22}
R.~A. Toni.
\newblock Theory on the dynamics of a parachute system undergoing its inflation process.
\newblock AIAA Paper 70-1170, 1970.

\bibitem[Huckins(1970)]{Ref23}
E.~K. Huckins.
\newblock Techniques for selection and analysis of parachute deployment systems.
\newblock NASA TN D-5619, 1970.

\bibitem[Huckins(1971)]{Ref24}
E.~K. Huckins.
\newblock Snatch force during lines-first deployment of space vehicles.
\newblock 8:\penalty0 298--299, 1971.

\bibitem[Purvis(1983)]{Ref25}
J.~W. Purvis.
\newblock Prediction of parachute line sail during lines-first deployment.
\newblock 20:\penalty0 940--945, 1983.

\bibitem[Eckroth et~al.(1993)Eckroth, Garrard, and Miller]{Ref26}
W.~V. Eckroth, W.~L. Garrard, and N.~Miller.
\newblock Design of a recovery system for a reentry vehicle.
\newblock AIAA Paper 93-1224, 1993.

\bibitem[French(1979)]{Ref27}
K.~E. French.
\newblock A first-order theory for the effects of line ties on parachute deployment.
\newblock AIAA Paper 79-0450, 1979.

\bibitem[Purvis(1984)]{Ref28}
J.~W. Purvis.
\newblock Improved prediction of parachute line sail during lines-first deployment.
\newblock AIAA Paper 84-0786, 1984.

\bibitem[Poole and Huckins~III(1972)]{Ref30}
L.~R. Poole and E.~K. Huckins~III.
\newblock Evaluation of massless-spring modeling of suspension-line elasticity during the parachute unfurling process.
\newblock NASA TN D-6671, 1972.

\bibitem[Whitlock and Poole(1973)]{Ref31}
C.~H. Whitlock and L.~R. Poole.
\newblock Postflight simulation of parachute deployment dynamics of viking qualification flight tests.
\newblock NASA TN D-7415, 1973.

\bibitem[McVey and Wolf(1974{\natexlab{b}})]{Ref32}
D.~F. McVey and D.~F. Wolf.
\newblock Analysis of deployment and inflation of large ribbon parachutes.
\newblock 11:\penalty0 28--33, 1974{\natexlab{b}}.

\bibitem[Heinrich(1973)]{Ref33}
E.~K. Heinrich.
\newblock A parachute snatch theory incorporating line disengagement impulses.
\newblock AIAA Paper 73-4641, 1973.

\bibitem[Johnson(1989)]{Ref34}
D.~W. Johnson.
\newblock Testing of a new recovery parachute system for the f-111 aircraft crew escape module: An update.
\newblock DE89-007139, 1989.

\bibitem[Maydew et~al.(1991)Maydew, Peterson, and Orlik-Rueckemann]{Ref35}
R.~C. Maydew, C.~W. Peterson, and K.~J. Orlik-Rueckemann.
\newblock Design and testing of high-performance parachutes.
\newblock AD-A246343, 1991.

\bibitem[Zhan et~al.(2014)Zhan, Yu, Yang, et~al.]{Ref36}
Yanan Zhan, Li~Yu, Xue Yang, et~al.
\newblock Initial stress correction method for the modeling of folded space inflatable structures.
\newblock 18:\penalty0 166--173, 2014.

\end{thebibliography}
\end{document}